\documentclass[10pt,twocolumn,letterpaper]{article}

\usepackage{cvpr}
\usepackage{times}
\usepackage{epsfig}
\usepackage{graphicx}
\usepackage{amsmath}
\usepackage{amssymb}
\usepackage{multirow}

\newcommand*{\bv}{\boldsymbol}

\usepackage[pagebackref=true,breaklinks=true,letterpaper=true,colorlinks,bookmarks=false]{hyperref}
\usepackage[breaklinks=true,bookmarks=false]{hyperref}
\cvprfinalcopy % *** Uncomment this line for the final submission
\def\cvprPaperID{****} % *** Enter the CVPR Paper ID here
\def\httilde{\mbox{\tt\raisebox{-.5ex}{\symbol{126}}}}

\begin{document}

%%%%%%%%% TITLE
\title{Sampled Image Tagging and Retrieval Methods on User Generated Content}

\cvprfinalcopy % *** Uncomment this line for the final submission
\def\cvprPaperID{****} % *** Enter the CVPR Paper ID here
\def\httilde{\mbox{\tt\raisebox{-.5ex}{\symbol{126}}}}

\author{Karl Ni, Kyle Zaragoza\\
Yonas Tesfaye, Alex Gude\\
\emph{Lab41, In-Q-Tel}\\
{\tt\small \{kni,kylez,ytesfaye,agude\}@iqt.org}
\and
Charles Foster \\
\emph{ Stanford University }\\
{\tt\small cfoster0@stanford.edu}
\and
Carmen Carrano, Barry Chen\\
\emph{ Lawrence Livermore }\\
\emph{ National Laboratory }\\
{\tt\small \{carrano2,chen52\}@llnl.gov}
}

\maketitle

%\thispagestyle{empty}
%%%%%%%%% ABSTRACT
\begin{abstract}
Traditional image tagging and retrieval algorithms have limited value as a result of being trained with heavily curated datasets. These limitations are most evident when arbitrary search words are used that do not intersect with training set labels. Weak labels from user generated content (UGC) found in the wild (e.g., Google Photos, FlickR, etc.) have an almost unlimited number of unique words in the metadata tags. Prior work on word embeddings successfully leveraged unstructured text with large vocabularies, and our proposed method seeks to apply similar cost functions to open source imagery. Specifically, we train a deep learning image tagging and retrieval system on large-scale, user generated content (UGC) using sampling methods and joint optimization of word embeddings. By using the Yahoo! FlickR Creative Commons (YFCC100M) dataset, such an approach builds robustness to common unstructured data issues that include but are not limited to irrelevant tags, misspellings, multiple languages, polysemy, and tag imbalance. As a result, the final proposed algorithm will not only yield comparable results to state of the art in conventional image tagging, but will enable new capability to train algorithms on large, scale unstructured text in the YFCC100M dataset and outperform cited work in zero-shot capability.
\end{abstract}

%%%%%%%%% BODY TEXT

%\pagebreak

\section{Introduction}

% Distinguish image tagging and retrieval versus classification
Automated approaches to tag and retrieve images, while distinct from image detection and localization, have benefited from some of the techniques developed for detection and localization competitions~\cite{ILSVRC,pascalVOC,mscoco} such as the rise of convolutional neural networks (CNNs) \cite{googlenet, vgg, AlexNet2012}. Such algorithms work well because the dataset is well-curated but are unfortunately limited by the number of keywords that can be used to tag and retrieve images due to a small training label set. To be useful, deep learning approaches need to accommodate open vocabulary search, which a practical implementation would require the extension of training label set to be orders of magnitude larger.

Extending curated datasets requires supervision, where developed algorithms would need to be tolerant of inevitable labeling errors. Conversely, open source imagery datasets from Google Photos or FlickR that are created with user generated content (UGC) have an almost unlimited number and variety of unique tags that cover much of the vocabulary of the English language. The problem with metadata tags in such datasets like the Visual Genome project~\cite{visgenome} and to a larger extent the YFCC100M dataset~\cite{yfcc100m} is that they also include noise in the form of misspellings, unevenly distributed numbers of tags (including auto-generated ones), different languages, irrelevant and unduly specific tags, just to name a few problems. Our approach to remedy the so-called ``weak labels'' in UGC is to statistically overcome the inherent noise with the sheer scale of the data. 

Leveraging UGC with data scale is non-trivial for a variety of reasons. While there is an abundance of work addressing \emph{image} scale in YFCC100M~\cite{karlpaper}, the focus of our work is on the scale of the \emph{labels}, which take the form of noisy metadata tags. 
The challenge then becomes negotiating matrix operations on any deep learning architecture that requires a final layer that is proportional to the number of words. In fact, the number of weights in that layer is (\#hidden units $\times$ \#unique tags), roughly 43M parameters in our case. As reported in~\cite{facebook}, the forward and backpropagation of a single batch on this final layer alone (a final layer that is $\frac{1}{4}$ the size of ours) takes 1.6 seconds, and certain heuristics must be employed. To put this in perspective, not considering the rest of the neural network, a single run through the YFCC100M dataset would take over two weeks.
% Put this in perspective.
% Make sure you figure out the numbers here.

% Fortunately, unstructured text and large vocabulary is well-studied with now-popular word embeddings~\cite{glove,word2vec}. Several works~\cite{fast0tag,0shot-hierarchical,0shot-embedding} have consequently sought to exploit word embeddings by projecting image features into the resulting semantic space. Such efforts primarily focus on zero-shot learning. By maintaining static word vectors, they assume that semantic and syntactic similarity necessarily equates to visual similarity, which is often not the case as word co-occurrence and parts of speech often have little to do with visual appearance. As a result, projecting into word embedding space merely \emph{uses} the word vectors and do not scale to nor truly train on unstructured UGC. 

Fortunately, unstructured text and large vocabulary is well-studied with now-popular word embeddings~\cite{glove,word2vec}. Several works~\cite{fast0tag,0shot-hierarchical,0shot-embedding} have consequently sought to exploit word embeddings by projecting image features into the resulting semantic space. Such efforts primarily focus on zero-shot learning. Targeting word vectors instead of tags means that the final layer size is no longer proportional to the number of words. By maintaining static word vectors, these approaches assume that semantic and syntactic similarity necessarily equates to visual similarity, which is often not the case as word co-occurrence and parts of speech often have little to do with visual appearance. As a result, projecting into word embedding space merely \emph{uses} the word vectors and do not scale to nor truly train on unstructured UGC.

The proposed method sidesteps this issue by jointly optimizing \emph{both} image and word embeddings, while simultaneously addressing the scale issue through the use of negative sampling and noise contrastive estimation~\cite{nce} that have pervaded the natural language processing community~\cite{word2vec}. Specifically, we use the traditional cross-entropy cost function and provide an analytical comparison to ranking cost functions~\cite{fast0tag,ranknet}, to which we also apply the proposed sampling methods for fair comparison. In doing so, we train against the largest UGC corpus currently available~\cite{yfcc100m}, and demonstrate that despite the issues, automated tagging using a sampled cost function can produce considerably more useful information than the original tags themselves. More importantly, we enable the capability that the user can search for an almost unlimited number of words and retrieve meaningful and relevant images.

\section{Related Work}

Methods for automatic image annotation have ranged from generative~\cite{generative-modeling,generative-modeling-2,generative-modeling-3} and discriminative models~\cite{svm-tagging} for image tags to nearest neighbor search-based approaches~\cite{Makadia2008,Guillaumin2009,metric-learning}. Helping to establish this line of research, Makadia et al.~\cite{Makadia2008} exceeded the state of the art with a simple $k$-nearest neighbor algorithm combined with a greedy approach to prioritization of tags from neighboring images. Subsequent work set the current standard for tag prediction performance using probabilistic models for tag generation that accounted for the nearness of an image’s neighbors and for the occurrence of rare tags. The aforementioned studies relied on a set of 15 global and local image features (\cite{fasttag,Guillaumin2009}; provided at {\footnotesize \verb"http://lear.inrialpes.fr/people/guillaumin/data.php"}) that have formed the basis of subsequent tag prediction research. As datasets grew larger, these non-parametric approaches became untenable despite producing the best quantitative retrieval results because complexity intimately relied on the number of images in the training set, deferring to dual projection methods~\cite{fasttag}.

% Focus on features
Subsequently, researchers began to focus more attention on the features themselves. Feature selection~\cite{metric-learning} on manual features through metric learning provided an adequate analysis of which manual features provided the most amount of information. Given their success in image classification~\cite{ILSVRC} and other domains~\cite{zhou2014}, investigations into combinations of features started to include deeply learning features~\cite{vgg,googlenet,AlexNet2012}. This prompted conflicting studies, where findings~\cite{mayhew-conflict} initially favored heavily tuned manual features. Successive and more thorough/complete studies~\cite{mayhew} demonstrated that deep features not only provided value to feature combinations, but can often outperform traditional methods on their own. %For the annotation problem at least, the conclusion on the predictive performance of these features in conjunction with and in comparison to manual features was that they here to stay. 
The conclusions have given rise to demonstrations of the predictive performance in the image tagging problem, where Kiros and Szepesvari~\cite{Kiros2012x} learned binary codes with real-valued image features derived from an unsupervised, autoencoder-based approach. The experiments presented in this work opt for pre-trained features\footnote{We did, indeed, do the full backpropagation through the entire neural network, to little benefit.} with our best results coming from \emph{Inception} features~\cite{googlenet}, though we are feature-agnostic and can use any fixed-dimensional vector representation.

% Data Tagging
Until recently, the related work in image tagging and retrieval has centered around earlier, smaller datasets~\cite{iaprtc12,espgame} (approximately 20k images, 300 unique tags) to recently released larger datasets~\cite{visgenome,nuswide} (over 100k/200k images, 5k object tags). The fact that the maximum vocabulary size of these curated datasets is on the order of thousands is of particular note since the Oxford English Dictionary places the count of unique, non-domain specific words in common use at roughly 170k.\footnote{There are 615k definitions (which include plural forms, different conjugations, and other derivatives of the original word), and the number of total words in use in the English Language is closer to a million.} Moreover, these annotations have an overwhelming bias towards objects rather than specific word concepts like actions, numbers, and more abstract ideas like color. 

% Zero-shot learning
To accommodate for training vocabulary shortcomings, zero-shot learning, initially made popular through Socher's recursive learning efforts \cite{socher-0shot}, provided an alternative by using the power of semantic information through word embeddings~\cite{word2vec,glove}. Later efforts typically involved some type of joint embedding~\cite{0shot-embedding,fast0tag}. In particular, Fast0Tag~\cite{fast0tag} identifies a principal direction in the word embedding space to which to project an image feature.

Instead of zero-shot learning, the proposed algorithm focuses its energy on one of the basic mantras of deep learning: that the proper way of overcoming deficits in capability is to introduce more data and add more parameters. To do so, our approach makes explicit use of the open source YFCC100M dataset~\cite{yfcc100m}. In contrast to the vocabulary in traditional tagging datasets, the metadata in the YFCC100M data is seemingly unbounded, where the number of unique tokens in the metadata tag exceeds 6M words over a variety of ideas. 

The earliest work on unstructured imagery in the YFCC100M corpus using deep learning~\cite{karlpaper} remained largely unsupervised and sidestepped the vocabulary scale issue. Other algorithms in an effort led by Facebook~\cite{facebook} followed with limited success by applying supervised approaches to the weak labels that the metadata tags provide. The bottleneck, it appeared, was the ability of the algorithms to train the final layer of the deep learning architecture on unstructured text in a reasonable amount of time. Alternatively, the domain that has been particularly successful in approaching the large vocabulary and unstructured text problems are word vector optimizations like \emph{Glove}~\cite{glove} and \emph{word2vec}~\cite{word2vec}. The proposed method leverages the sampling methods introduced in \emph{word2vec}, while comparing a few cost functions. In contrast to approaches that merely project image features into semantic space~\cite{fast0tag,0shot-embedding}, we leverage the embedding cost function itself, using a neural network that simultaneously optimizes over both words and images to produce a single vector space that can represent both modalities.

\section{Approach}

Typical deep learning image classifiers target a single concept with a \emph{softmax} layer in a neural network. This normalizes to create a probability distribution that expects a single label for a given image. While the single label assumption works for ImageNet's iconic images, it does not apply to images in the wild where they are coincident with several (probably noisy) tags. 

\begin{figure}[htbp]
  \centering
  \centerline{\includegraphics[width=8.5cm]{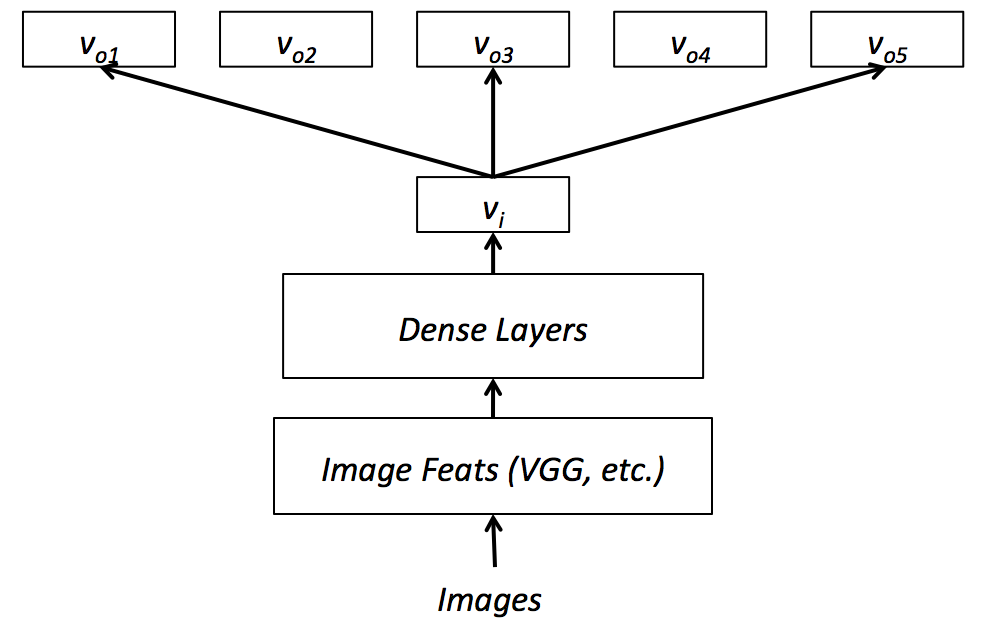}}
\caption{Training images and tags using ``skip-image'' optimization.}
\label{fig:im2vec}
\end{figure}

Instead, we turn to un-normalized cost functions observed by natural language processing and use optimization strategies rooted in unsupervised and embedding approaches. Most notably, Restricted Boltzmann Machines and word embeddings like \emph{word2vec} rely on some variant of noise contrastive estimation, where the distribution of foreground (e.g., the surrounding context of a word) is separated from the distribution of the background (e.g., the probability distribution over all words in the corpora). In our case, the context is the set of tags for each image, and negative samples can be obtained by sampling from the tag distribution. It is then straightforward use the skip-gram approach ( Fig.~\ref{fig:im2vec}) with the following cost function:

\begin{multline}
\mathcal{L}( \{W_i\}, \bv{v}_{p,n} ) = \sum_p^P \log \mathbb{E}_p\left[ \sigma (\bv{f}_{\{W_i\}}^T \bv{v}_p ) \right] + \\ \sum_n^N \log \mathbb{E}_n\left[ \sigma (- \bv{f}_{\{W_i\}}^T \bv{v}_n ) \right] + \\ \alpha \sum_{p,p'} \sigma (\bv{v}^T_{p} \bv{v}_{p'}) + \sum_{p,n} \sigma (-\bv{v}^T_{p} \bv{v}_{n})\label{eq:optfxn},
\end{multline}
where $\bv{v}_p$, are positively sampled vectors coming from words the image has been tagged with, $\bv{v}_n$ are the negatively sampled vectors from the probability distribution over all possible tags, $P$ is the number of positive samples, $N$ is the number of negative samples, and the feature vector $\bv{f}$ is parameterized by the set of weights $\{W_i\}$ from a neural network $h(\cdot, \{W_i\})$:
\begin{equation}
\bv{f}_{\{W_i\}} = h(\bv{x},\{W_i\})
\end{equation}
with inputs being ILSVRC features $\bv{x}$.

Like \emph{word2vec}, the first term in~\eqref{eq:optfxn} positively correlates the feature vector with the metadata tags, pulling the image closer to the context through backpropagation over $\{W_i\}$. The second term pushes the them away from the background distribution. The final two terms, with $\alpha$ taken to be small (we use 0.01), serves to promote similarity in co-occurring tags. These final terms are only useful when we are optimizing over $\bv{v}$, of which we provide an analysis later. Fig.~\ref{fig:t-SNE} depicts the vector projection of both images ($\bv{f}$) and text ($\bv{v}$) into the same joint vector space using t-SNE dimensionality reduction. 

The architecture of our neural network is very similar to that of \cite{fast0tag}, with hidden units of
\begin{equation}
\text{imfeats}\rightarrow4096\rightarrow8192\rightarrow2048\rightarrow300\rightarrow \text{wordvecs}, \nonumber
\end{equation} 
and we try a variety of \emph{Inception} and \emph{VGG} features with a combination of word vectors (\emph{word2vec} and \emph{Glove}). Take note that~\eqref{eq:optfxn} not only takes the set of weights $\{W_i\}$ from the neural network $h(\cdot)$ for parameters, but a key difference between the proposed method and most cited work~\cite{0shot-hierarchical,0shot-embedding,fast0tag} is that it is also a function of $\bv{v}_{p,n}$. That is, we are also optimizing the word vectors provided we have a large enough vocabulary. 

This makes sense when we note that word embeddings have a tendency to cluster semantically similar concepts, but these concepts sometimes are not visually similar. For example, every word embedding model that we have trained using both \emph{Glove} and \emph{word2vec} almost universally place words like ``red'',``green'', and ``yellow'' very close together, which makes sense semantically. Their context is the same, because one typically uses colors interchangeably, e.g., ``I have a (red/yellow) car.'' While semantically similar, these words are visually discriminating. For example, stop signs and green lights portray very different meanings despite their defining attribute being color. The same can be said of many other words, including numbers and titles of persons. Therefore, the word vectors $\bv{v}_p$ in~\eqref{eq:optfxn} for such instances should be somewhat separate from each other so that related image projections $\bv{f}$ can remain visually distinct.

\begin{figure}[htbp]
  \centering
  \centerline{\includegraphics[width=6.5cm, height=2.5cm]{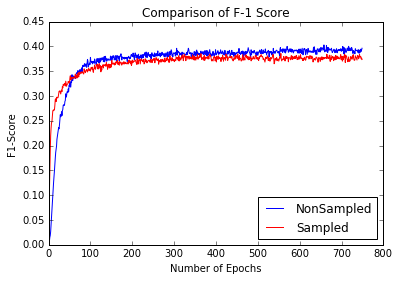}}
  \centerline{\includegraphics[width=6.5cm, height=2.5cm]{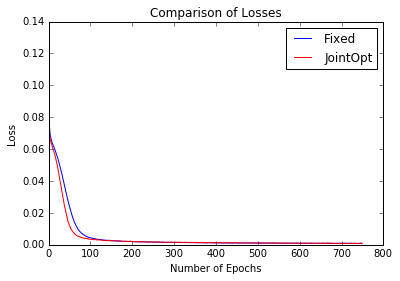}}
\caption{Top: Comparison in F1 optimization in the presence or absence of sampling. Bottom: Comparison in objective function optimization with and without joint optimization.}
\label{fig:sampling-loss}
\end{figure}

Inherent in~\eqref{eq:optfxn} is the idea that positive and negative sampling can converge to a meaningful result in expectation. A similar approach was initially attempted~\cite{fasttag} through single samples, though quickly abandoned due to updates being too sparse. In practice, with proper initalization of the final layer (i.e., using pre-trained word vectors) with a large corpus (we used New York Times and Wikipedia~\cite{text8b}), reasonable image retrieval results begin appearing and converging near a single pass through the YFCC100M data, at least for frequently occurring words. To assess the efficiency of sampling, see Fig.~\ref{fig:sampling-loss} which was created using a smaller corpora where it is possible to use traditional optimization in the cross-entropy loss function as a comparison point with respect to sampled loss.

Because we use deep learning to optimize~\eqref{eq:optfxn}, our variables are tensors and we require the number of samples to be consistent between images for batch purposes. That is, we must fix $P$ and $N$. Unfortunately, the image metadata have variable numbers of tags, and we choose a fixed number of positive samples, chosen uniformly since we make no assumptions on tag ordering. Although 69\% of images in YFCC100M have an average of 7.07 tags, we found that $P=5$ tags per image works well.
\label{sec:approach}

\subsection{On the Cost Function}

% Word2Vec is essentially sampled cross entropy
The proposed objective~\eqref{eq:optfxn} stands in stark contrast to prior work based on projections of images into semantic space based on ranking. Most notably, the Fast0Tag objective which originates from RankNet can be rewritten to take the form:
\begin{equation}
    \mathcal{L} = \beta \sum_{p} \sum_{n} \log \sigma\left( \bv{f}^T ( \bv{v}_p - \bv{v}_n ) \right) \label{eq:fast0tag}
\end{equation}
At first glance, the form of~\eqref{eq:fast0tag} appears quite different from~\eqref{eq:optfxn}. With some linear algebra and some simplifying assumptions, the contrasts are not as great as one might assume. In this section, we quantify the differences, and we highlight the pitfalls with respect to complexity and accuracy. On the latter point, ~\eqref{eq:optfxn} can be shown to be analogous to~\eqref{eq:fast0tag} but for two additional terms.

Let us ignore the last two terms in~\eqref{eq:optfxn} for now, as they simply tweak performance. It can be shown that comparisons with the first two terms of the proposed objective turn out to be a sampled cross-entropy function. Therefore, the word vectors are simply the last layer on a traditional neural network where only a few columns actually matter. As discussed, we are sampling $P$ and $N$, but for sake of explanation, let us assume that $P=N$ (it does not), remove the expectations, and for clarity, let's assign $L=P=N$. Then,
\begin{eqnarray}
    \mathcal{L} &=& \sum_p^L \log \sigma (\bv{v}_p^T \bv{f}_n ) + \sum_n^L \log \sigma ( -\bv{v}_n^T \bv{f} ) \nonumber \\
    &=& \sum_p^L \sum_n^L \frac{1}{L} \log \sigma (\bv{f}^T\bv{v}_p) + \frac{1}{L} \log \sigma (-\bv{f}^T\bv{v}_n) \nonumber \\
    &=& \sum_p^L \sum_n^L \frac{1}{L} \log \left( \sigma (\bv{f}^T\bv{v}_p ) \sigma( -\bv{f}^T\bv{v}_n) \right) \nonumber \\
    &=& \beta' \sum_p^L \sum_n^L \log \left( \sigma\left( \bv{f}^T (\bv{v}_p - \bv{v}_n) \right) +  e^{\bv{v}_p^T \bv{f}} + e^{-\bv{v}_n^T \bv{f}} \right) \nonumber \\ \label{eq:derive}
\end{eqnarray}

Besides scaling by $\beta'=\frac{1}{L}$, the equivalence shown in \eqref{eq:derive} is directly comparable because the first term inside the logarithm mirrors~\eqref{eq:fast0tag}. Conceptually, the salient point arises from the idea that  \eqref{eq:fast0tag} ranks the \emph{difference} between every positive example to every other negative example. This inherently maps general vector directionality rather than the actual correlation between the vectors themselves. Meanwhile, the extra terms in~\eqref{eq:derive} are dot products between images and words to maximize or minimize. The effect is essentially to sum nonlinear correlations between individual images and its relevant individual tags. As is evident in Table~\ref{conventional} in the results sections, it is these extra terms that lead to an improvement in performance during conventional tagging. Secondly, such an effort will also enable the objective function to improve with fewer constraints, meaning that higher accuracy and quicker training times can be achieved as seen in Fig.~\ref{fig:sampling-loss} and Table.~\ref{timing}. We compare both objective functions with and without optimization and sampling in Sec.~\ref{sec:results}.

% \begin{figure}[htbp]
%   \centering
%   \centerline{\includegraphics[width=8.5cm]{images/Curves.png}}
% \caption{Comparison of timing information between algorithms.}
% \label{fig:opt}
% \end{figure}
\begin{table}[h!]
    \begin{center}
     \begin{tabular}{|c | c c | c c | } 
     \hline
     \multicolumn{5}{|c|}{Timing Per Epoch} \\
     \hline
     & \multicolumn{2}{|c|}{291 Labels} & \multicolumn{2}{|c|}{925 Labels} \\
     & \multicolumn{2}{|c|}{17665 Images} & \multicolumn{2}{|c|}{102709 Images} \\
     \hline     
     & Full & Sampled & Full & Sampled \\
     \hline
     Fast0Tag~\cite{fast0tag}      & 2.12s & 0.39s & 43.6s &  3.22s \\
     X-Ent                         & 1.21s & 0.32s & 9.3s & 1.94s  \\
     Opt+XE                        & 1.30s & 0.43s & 10.9 & 2.31s \\
     \hline
    \end{tabular}
    \caption{Timing Information.}
    \label{timing}
    \end{center}
\end{table}

Also, we are explaining away the non-trivial double sum to make a point, but its presence is quite significant when considering computational complexity. In reality, \eqref{eq:optfxn} is implemented with computational complexity $\mathcal{O}(\max(P,N))$ per image. Because of the double sum over $p$ and $n$, fast0tag has complexity (and memory since we're using tensor implementations) that scales according to $\mathcal{O}(PN)$ per image. Without sampling, if there are more than a few labels per image, this balloons quite significantly, particularly if there are large numbers of positive labels for a single image. The issue is that the optimization is essentially ranking every tag to every other tag for every single image in a minibatch. In the absence of sampling, cross entropy turns out to be much faster. In our experiments, transitioning to larger datasets using full optimization (without sampling) from 291 tags (IAPR TC-12~\cite{iaprtc12} 2s/epoch) to 925 tags (NUS-WIDE~\cite{nuswide}, 43s/epoch) was significant, and from 925 tags to 13980 (Visual Genome~\cite{visgenome}) was untenable on our TitanX GeForce NVIDIA card due to both memory and time.

\subsection{Out of Vocabulary Updates}
\label{sec:outofvocab}

While the tag set in a UGC corpus like YFCC100M has a surprisingly extensive coverage over the entire English language, it still does not include all possible words that can be used for image search. As stated previously, we pre-train on a text corpus, e.g. Wikipedia's first 8 billion characters~\cite{text8b}, but the set difference between the word corpus and the metadata in the image corpus is non-trivial. Because we only train on samples in the image corpus, words in the set difference will never be updated when jointly optimizing in Sec.~\ref{sec:jointopt}. The relationship between optimized and unoptimized words quickly devolves, and the dot product between optimized word vectors and un-optimized word vectors becomes meaningless.

\begin{figure}[htbp]
  \centering
  \centerline{\includegraphics[width=8.5cm]{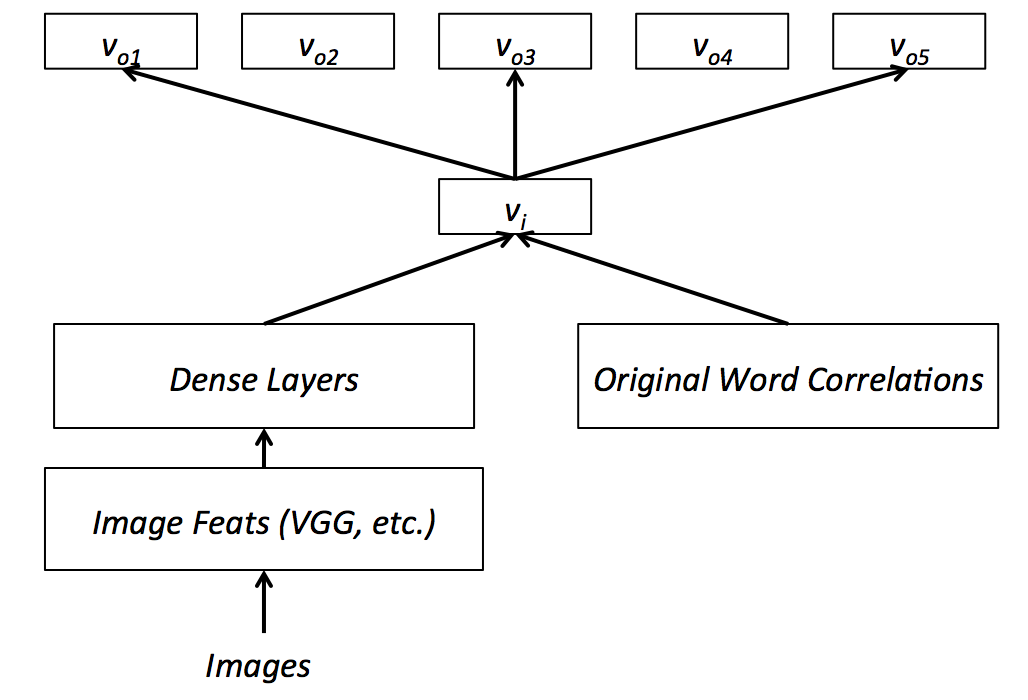}}
\caption{Jointly optimizing word vectors and image weights.}
\label{fig:sample-words}
\end{figure}

The approach that we take in Fig.~\ref{fig:sample-words} is to simply make a final optimization pass after training the neural network on the image corpus. Note that this step makes the most sense when performed \emph{offline}, simply snapping out-of-vocabulary words into place. Assuming we have trained a word embedding beforehand, we can then apply what we know of the semantic relationship between the out-of-vocabulary words and the in-vocabulary words, i.e., words that have been optimized when training the neural network.

Let $V_{W}$ be word vectors from the text corpus (e.g., Google News), and $V_{I}$ be word vectors from the image corpus (e.g., YFCC100M). The set difference is thus $\{V_D\} = \{V_W\} - \{V_I\}$. Without loss of generality, let us assume that all words in the image corpus exist in the word corpus, i.e. $\{V_I\} \subset \{V_W\}$ and furthermore, we can organize $V_\cup$ to be ordered in the following manner:
\begin{equation}
    V_\cup = \left[ V_I | V_D \right] \nonumber
\end{equation}
Before training in Sec.~\ref{sec:approach}, let us preserve the initial word vectors $V_I$ and $V_D$. Then for the sake of argument, we can write a nonlinear correlation matrix\footnote{Note that for computational purposes, we do not explicitly construct $C^{(i)}$, but rather store a subset of the original word vectors and compute dot products online.} for every word vector to every other word vector as:
\begin{equation}
C^{(i)} = \sigma \left(
\begin{bmatrix}
V_I^T V_I & V_I^T V_D \\
V_D^T V_I & V_D^T V_D
\end{bmatrix}
\right)
\label{eq:corrmat},
\end{equation}
where the superscript $i$ denotes the initial semantic relationships of each word to one another. After training with our image corpus, all the vectors in $V_I$ will have changed while none of the vectors in $V_D$ will have been updated. %That means the upper left-hand corner of the final correlation matrix $C^{(f)}$ will look very different, and the remainder of $C^{(f)}$ prior requires additional tuning.
%
% Let $V^{(f)}_I$ be the final tuned word vectors in the image corpus, where we rely on~\eqref{eq:optfxn} to have updated each column. We fix these vectors. It is the remainder of $C^{(f)}$ requires some additional tuning.
%
% If we say that this upper left-hand corner of $C^{(f)}$ is already optimally tuned from~\eqref{eq:optfxn}, then the only free parameters left to optimize are $V_D$. 
To update $V_D$ in the absence of any image information, we can only rely on semantic information, which are specified by the relationships in $C^{(i)}$ the lower and right-hand submatrices of $C^{(i)}$. 
% That means that the lower and right-hand submatrices in $C^{(f)}$ must satisfy the same relationships as specified by $C^{(i)}$. 
Specifically, we wish to match initial and final correlations from seen words to unseen words and between the unseen words themselves:
% , matching the initial and final submatrices of $C_{ID}$, $C_{DI}$, and $C_{DD}$.
\begin{equation}
C^{(i)}_{d,m} \log \sigma( \bv{v}_d^T \bv{v}_m ) + (1 - C^{(i)}) \log\left( 1 - \sigma(\bv{v}_d^T\bv{v}_m ) \right)
\end{equation}
for $\bv{v}_d \in \{V_D\}$ and $\bv{v}_m \in \{V_\cup\}$. 

% Essentially, the goal is to make the non-optimized portions of $C^{(f)}$ be very similar to the corresponding entries in $C^{(i)}$. In our case, the image corpus has enough words so that $V_D$ are often overdetermined. However, we require a little thought be put into

\section{Implementation Issues}
\label{sec:jointopt}

To replicate our work (code at \verb"http://github.com/lab41/attalos"), nontrivial details remain in the execution of the ideas in Sec.~\ref{sec:approach} due to the scale of the data. Such considerations include pre-training the final layer of our neural network, the partition of the architecture onto GPU/CPU memory, and additional tricks that are necessary for sampling. 

\subsection{Pre-training the Final Layer}

It is worth noting that even with 100M images, less popular concepts may occur with relatively low frequency. Meanwhile, despite the data being sufficiently ``big'', the number of parameters in the final layer of our neural network architecture is quite considerable. As the vocabulary is still somewhat proportional to the number of images, pre-training will alleviate the concerns that occur when initializing from scratch. Because each column of the final weighting matrix is a vector corresponding to a unique word, we pre-train this final layer with the a word corpus using \emph{word2vec} or \emph{Glove}\footnote{We tried both; \emph{Glove} tended to provide better results, though we did not explore too many text corpora.}. We pre-trained using several corpora that included Google News, New York Times, and settled on Wikipedia 8B. The dimensionality of the hidden layer before the final word embedding is 300, and thus, each word vector is 300 dimensions, and this is stored in a $432213 \times 300$ weight matrix.

\subsection{GPU/CPU Split}

% Without pruning, the number of tags in UGC can number in the tens of millions without constraints on language or spelling errors. On this note, 
We would like to keep the algorithm and parameters in memory in order to achieve quick training times, though we note that the latest computer vision algorithms have made exclusive use of the GPU. Work in deep learning suggest that negotiating GPU DRAM (memory) hampers the flexibility that a researcher needs to fully explore a domain~\cite{gpu-memory}. In all actuality, GPU memory capacity and sharing have made remarkable advances in hardware. At least for now though, servers with NVLink and multi-GPU nodes are expensive, onboard server memory is ever more abundant, and so look to exploit onboard server memory, instead. 

Mikolov's original \emph{word2vec} is trained using multi-threaded CPUs. Since his work deals primarily with unstructured text, his neural network is wide rather than deep, using a single hidden layer. For this reason, our algorithm makes use of both CPU and GPU. We place the word embeddings (final layer of the neural network) off GPU, and the corresponding gradient calculations in~\eqref{eq:optfxn} are done with the CPU. The backpropagation through the remainder of the neural network (all the dense layers prior) are still done on the GPU using Tensorflow. Such a split makes training times quicker and feasible. % The deep learning component needs only be of the dimension of the word embedding (i.e., 300, in our case.) This is, as opposed to having to transfer a tensor with a dimension equal to the vocabulary size (432k). The latency is negligible, and almost preferable. This keeps all word vectors resident in memory, which in our case, is considerably larger than that of the GPU. 

\subsection{Scaled Sampling}

Like~\cite{word2vec}, we use words that occur with the highest frequencies and cut off the number of words at 423k. To ensure consistent tensor dimensions, for each image, we sampled 5-10 positive tags uniformly.~\footnote{In future implementations, we imagine this would be some form of an inverse distribution of the tag frequency.} Unfortunately, sampling negative examples from a 432k dimensional distribution is time consuming. In fact, our original code profile assessed that 60\% of time was spent in sampling.

To side-step this issue, we explored two options: (1) pre-sampling at each epoch and (2) using the metadata from an adjacent image. The first option is done by taking a random subset from the distribution of tags in the corpus at each epoch. Then during backpropagation, we sample again from this random subset. This sort of fast sampling may create inherent bias issues as the chances of re-sampling frequently used words are likely. Because the images are randomly ordered and selected at each epoch, the second option, to use the metadata from the next image in the epoch as a negative sample avoids this problem. Doing so reduced computation time to negligible rates.

Overall, by sampling, we can iterate through an entire epoch of YFCC100M in under three hours. The bulk of our models converged to meaningful results within a \emph{single} epoch. What follows in Sec.~\ref{sec:results} were rapidly prototyped for 40 epochs through the YFCC dataset.

 \begin{table*}[]
    \begin{center}
     \begin{tabular}{| c | c c c | c c c | c c c | c c c | c c c | } 
     \hline
     & \multicolumn{6}{|c|}{NUS-Wide Dataset~\cite{nuswide}} & \multicolumn{9}{|c|}{Multi-Corpora} \\ 
     \hline
     & \multicolumn{3}{|c|}{925$\rightarrow$81}
     & \multicolumn{3}{|c|}{925$\rightarrow$1006} & \multicolumn{3}{|c|}{IA$\rightarrow$ESP} &
     \multicolumn{3}{|c|}{VG$\rightarrow$ESP} &
     \multicolumn{3}{|c|}{YFCC$\rightarrow$ESP} \\
     \hline     
      & P & R & F1 & P & R & F1
      & P & R & F1 & P & R & F1 
      & P & R & F1 \\ [0.5ex] 
     \hline\hline
     Fast0Tag & 16.2 & \textbf{39.3} & 22.9 & 15.6 & 14.9 & 15.2 & 15.7 & 6.9 & 9.6 & \multicolumn{3}{|c|}{ Will Not Scale } & \multicolumn{3}{|c|}{\multirow{3}{*}{ Will Not Scale }} \\
     X-Ent       & \textbf{21.2} & 36.3 & 26.9 & 17.1 & 14.8 & 16.3 & \textbf{16.2} & 6.1 & 9.4 & 14.1 & 17.9 & 15.7 & & &   \\
     Opt+X-Ent   & 21.1 & 37.3 & \textbf{27.0} & \textbf{17.3} & \textbf{15.9} & \textbf{16.1} & 16.3 & \textbf{7.0} & \textbf{9.8} & \textbf{14.8} & \textbf{18.1} & \textbf{16.3} & & & \\
     \hline
     & \multicolumn{15}{|c|}{Sampled Methods} 
     \\
     \hline
     S+Fast0Tag & 15.0 & 43.4 & 22.3 & 12.4 & 10.3 & 11.6 & 5.9 & 8.3 & 6.9 & \textbf{17.6} & 18.4 & 18.0 & 5.1 & 3.9 & 4.4 \\
     S+X-Ent  & \textbf{15.9} & 44.2 & \textbf{22.9} & 13.0 & 11.3 & 12.1 & 10.0 & 5.2 & 7.1 & 16.6 & \textbf{19.8} & 18.1 & 17.4 & \textbf{17.6} & 17.5 \\
     S+Opt+X-Ent  & 15.4 & \textbf{44.6} & 22.9 & \textbf{13.1} & \textbf{11.6} & \textbf{12.3} & \textbf{13.3} & \textbf{10.2} & \textbf{12.1} & 17.3 & 19.0 & \textbf{18.1} & \textbf{21.9} & 15.1 & \textbf{17.9} \\
     [1ex] 
     \hline
    \end{tabular}
    \caption{Zero-shot and multi-corpus tagging top 5 results for precision/recall/F1. Due to space constraints, we report at limited precision, but the bolded results are the hightest results at full precision.}
    \label{generalization}
    \end{center}
\end{table*}

\begin{figure*}[htbp]
  \centering
  \centerline{\includegraphics[width=17cm]{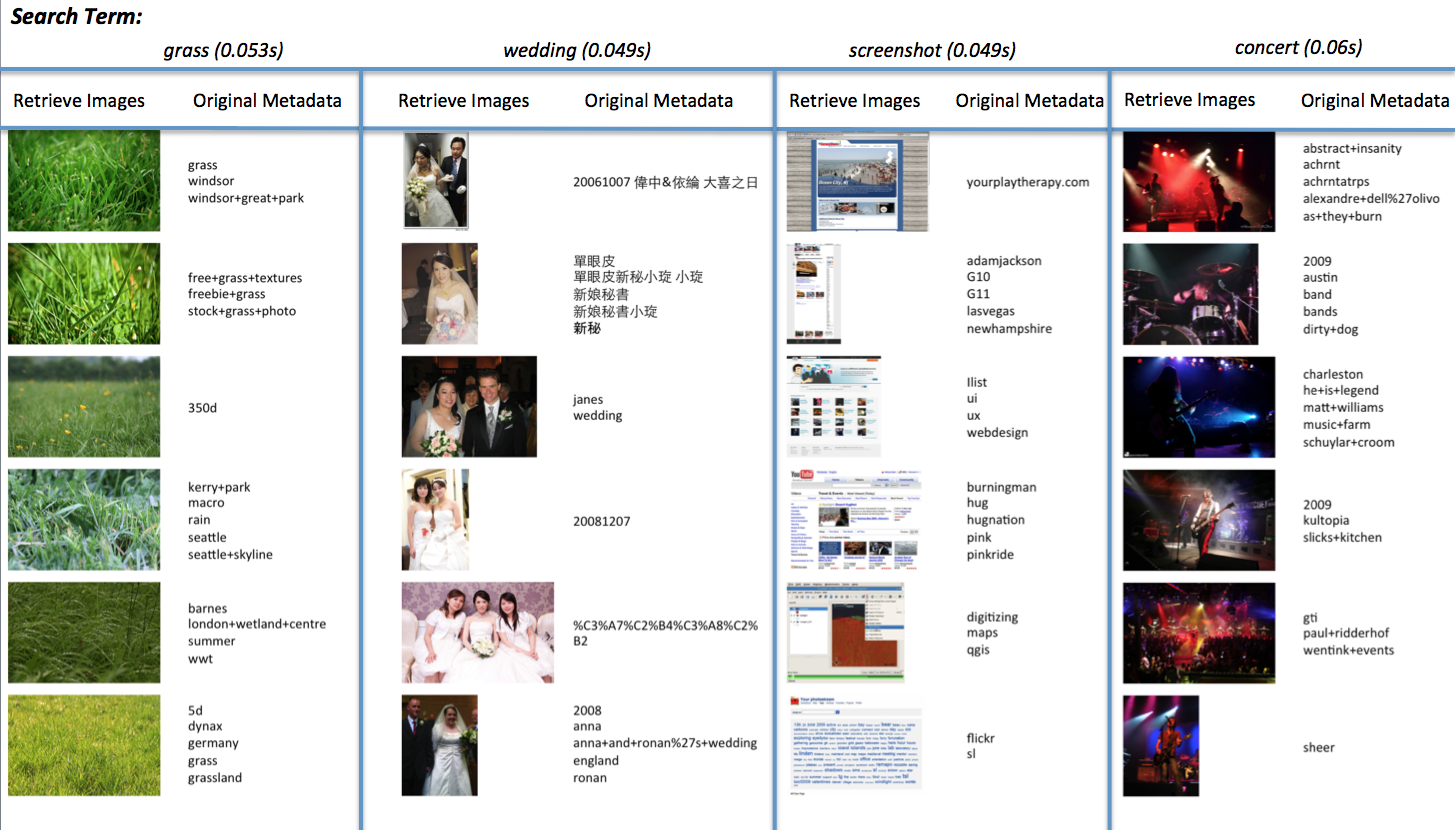}}
\caption{YFCC Image Retrieval and Truth Tags. Each pair of columns is a search term, and the top six images are retrieved. To demonstrate how much we rely on the statistical properties of the dataset, the metadata tags are also shown next to each image retrieved. In many cases, the metadata is in a different language (which we had to convert from URL encoded strings in UTF-8.) On a side note, it is apparent that a significant portion of weddings in the YFCC100M occur in Taiwan.}
\label{fig:YFCC}
\end{figure*}

\section{Results}
\label{sec:results}

While the primary objective is to train on UGC, we perform quantitative metrics on cured, traditional corpora in order to compare against state of the art. These include IAPR TC-12 (IA - 291 unique tags)~\cite{iaprtc12}, ESP-game (ESP - 288 tags, INRIA-LEAR's version)~\cite{espgame}, NUS-WIDE (using splits from~\cite{fast0tag}, 81/925/1006 tags)~\cite{nuswide}, Visual Genome (VG 13980 unique tags) ~\cite{visgenome}, and YFCC100M~\cite{yfcc100m} (millions, but we pruned to 432k tags) with both InceptionNet and downloaded YFCC-VGG features~\cite{yfcc-feats}. 

To assess image tagging capability, we train, validate, and test against proper splits from a single corpus. To assess generalization capability for a variety of content and word tags, we perform cross-corpus evaluation: train on a single dataset, then test on a totally different dataset. For example, IA$\rightarrow$ESP is trained and validated on the IAPR TC-12 training/validation splits and tested on the ESP-game test split. For YFCC100M, we only test on ESP-Game since YFCC100M$\rightarrow$YFCC100M evaluation is not meaningful due to noisy truth data. Along with cited algorithms in Table~\ref{conventional}, neural network approaches (both sampled and unsampled) using the cross-entropy function (with joint optimization) have run for a total of 750 epochs of the data. The effect of sampling on accuracy can be shown to be negligible during training while vastly improving running times as previously shown in Table~\ref{timing}.

\begin{figure*}[htbp]
  \centering
  \centerline{\includegraphics[width=18cm, height=7.3cm]{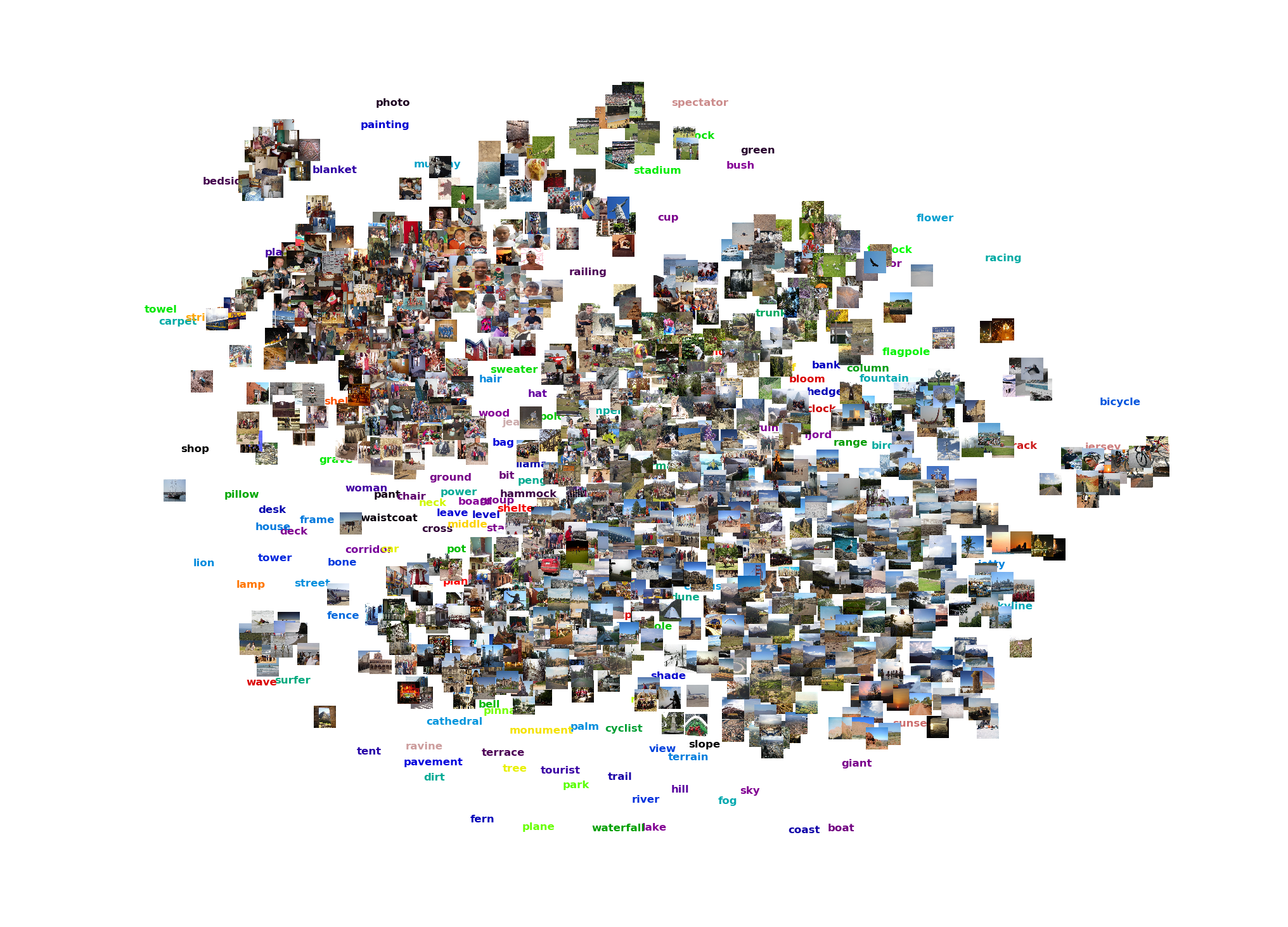}}
\caption{Qualitative example of jointly optimized vector space embedding, where both images and text live in the vector space.}
\label{fig:t-SNE}
\end{figure*}

The proposed methods are denoted in Table~\ref{generalization} and Table~\ref{conventional} as X-Ent, Opt+X-Ent, and S+Opt+X-Ent. The ``Opt'' stands for an optimized final word vector layer, the ``S'' stands for the sampled version, and again, X-Ent essentially means a cross-entropy cost function with a word vector final layer. We also include ``AvgWV'' as a benchmark in Table~\ref{conventional}, referring to an approach using the average of an image's tag vectors as a target for the deep neural network.

We use provided implementations of Fast0Tag~\cite{fast0tag} from the author as well as an original implementation using sampling, denoted in the tables as S+Fast0Tag. Such an implementation was able to achieve reasonable runtimes (on par with cross-entropy) that plagued the original implementation, which is evident in Table~\ref{timing} in Sec.~\ref{sec:approach}. We used the sampled code in place of the provided code for larger datasets, where the provided code was incapable of performing due to memory and complexity issues. It is important to note that many of the algorithms compared against~\cite{Guillaumin2009,fasttag}, especially the nearest neighbor ones, simply cannot address the large scale vocabulary in UGC nor the quantity of images in a reasonable amount of time, either at inference time or during training time. 

\begin{table}[h!]
    \begin{center}
     \begin{tabular}{|c | c c c | c c c | } 
     \hline
     \multicolumn{7}{|c|}{Image Annotation} \\
     \hline
     & \multicolumn{3}{|c|}{IA~\cite{iaprtc12}} & \multicolumn{3}{|c|}{ESP~\cite{espgame}} \\
     \hline     
     Top 5 & P & R & F1 & P & R & F1 \\
     \hline\hline
     least-squares                 & 40 & 19 & 26 & 35 & 19 & 25 \\
     Avg WV                        & 21 & 13 & 16 & 38 & 20 & 26 \\ 
     TagProp~\cite{Guillaumin2009} & 45 & 34 & 39 & 39 & 27 & 32 \\
     FastTag~\cite{fasttag}        & \textbf{47} & 26 & 34 & \textbf{46} & 22 & 30 \\
     Fast0Tag~\cite{fast0tag}      & 41 & 33 & 36 & 38 & 35 & 36 \\
     X-Ent                      & 44 & 34 & 39 & 38 & 36 & 37 \\
     Opt+XE                      & 43 & 36 & \textbf{39} & 38 & 36 & 37 \\
     \hline
     & \multicolumn{6}{|c|}{Using Sampling}
     \\
     \hline
     S+Fast0Tag                    & 37 & 38 & 37 & 37 & 38 & 38 \\
     S+X-Ent                       & \textbf{38} & 39 & 38 & 37 & 37 & 37 \\
     S+Opt+XEnt                    & 38 & \textbf{39} & 38 & 37 & \textbf{39} & \textbf{38} \\
     [1ex] 
     \hline
    \end{tabular}
    \caption{On the effect of using Fast0Tag and other methods versus the cross-entropy cost function for a single corpus evaluation. We selected the higher number at precision.}
    \label{conventional}
    \end{center}
\end{table}

Across both Table~\ref{generalization} and Table~\ref{conventional}, optimized cross-entropy outperforms Fast0Tag~\cite{fast0tag}, whether sampled or non-sampled. Although the goal in sampling was not to achieve the highest accuracy but to deal with scale, in many cases, adding sampling to the approach boosted recall for both conventional and cross-corpora evaluation. Our explanation for this phenomenon is that sampling helped to deal with noisy tagging.

Zero-shot capability is baked into the NUS-WIDE dataset with splits developed by~\cite{fast0tag} and varying numbers of tags. The 925 tag split does not intersect with 81 tag splits, and the 1006 tag split is the union. We also include \emph{cross-corpora} results where we train on one dataset and test on another dataset, of which the tag sets overlap some. However, the goal of the proposed algorithm is generalization using a large vocabulary. This was achieved through two large datasets: to a lesser extent, the Visual Genome object dataset (a subset of annotations distinct from the localization and captioning set) and to a larger extent, the YFCC100M dataset. We perform some pre-processing, removing images without any tags, and cutting off low-frequency words to achieve a tag count of approximately 14k and 432k words, respectively.

We also assessed the qualitative merits of the approach. The joint vector space embedding of $\bv{f}$ and $\bv{v}$ from~\eqref{eq:optfxn} is shown in Fig.~\ref{fig:t-SNE}, where the means to search between image and text modalities (tagging vs retrieval) is conducted by finding the most similar vector via normalized dot product. Meanwhile, results for image retrieval in Fig.~\ref{fig:YFCC} are conducted on a held out set of YFCC100M alongside their true metadata to provide an idea of the noise in the dataset. Of particular note is that we chose six terms to search for at random, and most of the best images returned had meaningless truth tags, which speaks to the power of our approach. The exhaustive top $N$ search was conducted in about 0.05 seconds per query on the CPU over a set of roughly 1.7 million normalized featurized images, which were held out prior to training.  % Because we open source imagery, we also include the metadata to convey just how unstructured the dataset is. The images were chosen at random, but the query.

\section{Conclusions}

We have proposed a method that can tag and retrieve images with UGC scale vocabulary through joint image and word vector optimization and sampling methods. We have demonstrated that our tagging mechaniscm can yield considerably more useful information than the original tags themselves.

% BiBTeX files (here: refs). The IEEEbib.bst bibliography
\pagebreak
\section{Acknowledgements}

We would like to thank Yang Zhang~\cite{fast0tag} for his openness, conversations, and help in replicating results on \emph{fast0tag} and related algorithms. Additionally, our initial conversations with Christopher Re (Stanford University) were very helpful. Patrick Callier (Lab41/IQT) had thoughtful comments and contributed to the text. Sriram Chandresekar and Bob Gleichauf provided useful direction and made the work possible.

\bibliographystyle{ieee}
\bibliography{refs}

\end{document}

%%%%%%%%% TITLE
\title{Sampled Image Tagging and Retrieval Methods on User Generated Content \\ Paper Supplement}

\author{Karl Ni, Kyle Zaragoza\\
Yonas Tesfaye, Alex Gude\\
Lab41, In-Q-Tel\\
{\tt\small \{kni,kylez,ytesfaye,agude\}@iqt.org}
\and
Carmen Carrano, Barry Chen\\
Computational Engineering\\
Lawrence Livermore National Laboratory\\
{\tt\small \{carrano2,chen52\}@llnl.gov}
}

\maketitle
\appendix

Due to space constraints in our our original submission, we were not able to include several plots, tables, qualitative evaluations, and  derivations. This document supplements our original paper with the additional information. As the original work explores sampling on deep learning cost functions, the majority of this supplement is devoted to supporting material to that effect. Sec.~\ref{sampling} examines the effect of sweeping both positive and negative numbers of sampling, where we find that a large number of samples are not required for optimal performance. Additionally, we provide intuition and derivations about how we mathematically arrived at a sampled cross-entropy cost function. This is provided in Sec.~\ref{derivation}. Sec.~\ref{morequeries} has more qualitative examples of the software that we have developed. We also noticed some discrepancies between the evaluation metrics of cited papers from the original paper, and we outline the exact differences in the Sec.~\ref{eval}. Finally, we include some additional results in the last section.

\section{On the Effect of Sample Numbers}
\label{sampling}

\begin{figure}[h]
    \centering
    \begin{subfigure}[b]{0.44\textwidth}
        \centering
        \includegraphics[width=\textwidth]{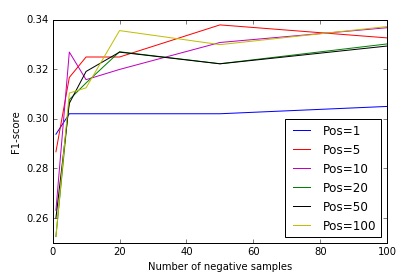}
        \caption{The relative comparisons of increasing the number of negative samples. (IAPR-TC12~\cite{iaprtc12})}
        \label{varyk}
    \end{subfigure}
    \hfill
    \begin{subfigure}[b]{0.44\textwidth}
        \centering
        \includegraphics[width=\textwidth]{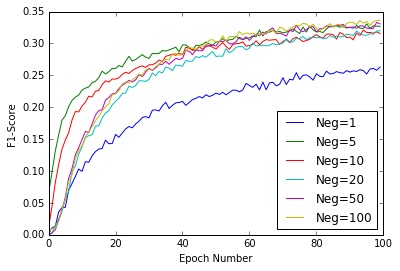}
        \caption{Convergence properties varying number of negative samples, positive samples fixed at 10. (IAPR-TC12~\cite{iaprtc12})}
        \label{fig:convergence}
    \end{subfigure}
    \hfill
    \caption{Varying the number of samples used.}
    \label{fig:sweep}
\end{figure}

The algorithm in the paper was chosen to use the number of positive samples at $P=5$ and the number of negative samples at $N=10$. We explored the number of samples used for positive and negative examples in depth, and discuss Fig.~\ref{fig:sweep} in detail in this section.

The knee in the curve appears in Fig.~\ref{varyk} at under 10 for positive and negative samples (which is why we chose $P=5$ and $N=10$). Another of the more concrete conclusions in Fig.~\ref{fig:sweep} is that single samples in both positive and negative labels universally lag in behind larger $P$ and $N$ across 100 epochs. In Fig.~\ref{varyk}, we observe that performance approaches near-parity when $P\ge5$ for larger numbers of samples. More surprising, though, is the fact that the number of negative samples seems to have less of an effect than previously thought as we observe similar performance when $N\leq100$ neg samples in Fig.~\ref{fig:convergence}. This has implications in how neural networks are trained in general when it is noted that training and optimizing for all the zeros in one/multi-hot encodings is common. The plots seem to suggest that this is unnecessary.

\section{The Sampled Cross-Entropy Cost Function}
\label{derivation}

The proposed algorithm is rooted in the cross-entropy function and the \emph{word2vec} model~\cite{word2vec}, and this section shows the equivalence of our sampled cost.

The cross-entropy loss for image $i$ can be written as:
\begin{multline}
    \mathcal{L}_i = \bv{y}_i^T \log \sigma( g(\bv{x}_i)) + \\
     (1 - \bv{y}_i^T) \log ( 1 - \sigma(g(\bv{x}_i) ) )
\end{multline}
where $g(x)$ is a neural network that takes in an image feature $\bv{x}_i$, and $\bv{y}_i \in \mathbb{R}^L$ are the output labels for image $i$. Let us separate the weight matrix of the last layer of the neural network, $V$, out from $g(\cdot)$ and call it $\bv{f}$ where $g(\cdot) = V \bv{f}$. As in the paper, $V$ can be thought of as the word matrix, and each row can be thought of as a word vector. In our case, we use word vectors of 300 dimensions, meaning $V \in \mathbb{R}^{L \times 300}$ and $\bv{f} \in \mathbb{R}^300$. Then, we write:
\begin{multline}
    \mathcal{L}_i = \bv{y}_i^T \log \sigma( V \bv{f} ) +
     (1 - \bv{y}_i^T) \log ( 1 - \sigma(V \bv{f} ) ) 
\end{multline}
In our formulation, let $S \in \mathbb{R}^{\text{\#samps} \times L}$ be a sampling matrix that is number of samples by $L$, where each column is one hot encoded to the sample number. For example, if we have total positive samples at $5$ and negative samples at $10$, then $S^p \in \mathbb{R}^{L \times 5}$ and $S^n \in \mathbb{R}^{L \times 10}$. Sampling both positive and negative examples, we have:
\begin{multline}
\mathcal{L} = ((S^p_{i}\bv{1}) \odot \bv{y}_i)^T \log \sigma(V \bv{f}) + \\
    ((S^n\bv{1}) \odot (1 - \bv{y}_i))^T \log ( 1 - \sigma(V \bv{f} ) )     
\end{multline}
where $\bv{1}$ is a vector of ones of appropriate size. Assuming that we are only sampling from the appropriate labels, we can drop $\bv{y}$ altogether, and $\bv{s}$ becomes a selector of sorts for the last layer of the neural network. Now, we can distribute $S^{(\cdot)}$ into the sigmoid function, selecting the appropriate columns of $V$:
\begin{equation}
\mathcal{L} = \bv{1}^T \log \sigma(S^{pT} V\bv{f}) +
    \bv{1}^T \log ( 1 - \sigma( S^{nT} V h(\bv{f}) ) )   
\end{equation}
Writing the sampling out as expectations, and using summation notation instead of using the $\bv{1}$ vector, we arrive at the same equation as the first two terms in (1) in the original paper:
\begin{eqnarray}
\mathcal{L} &=& \sum_p^P \log \mathbb{E}_p\left[ \sigma( \bv{v}_p^T \bv{f}) \right] + \nonumber \\
            && \qquad \qquad \sum_n^N \mathbb{E}_n\left[  \log ( 1 - \sigma( \bv{v}_n^T \bv{f}) ) \right] \nonumber \\
            &=& \sum_p^P \log \mathbb{E}_p\left[ \sigma( \bv{v}_p^T \bv{f}) \right] + \nonumber \\
            && \qquad \qquad \sum_n^N \mathbb{E}_n\left[  \log ( \sigma( -\bv{v}_n^T \bv{f}) ) \right]
\end{eqnarray}

\section{Additional Image Queries}
\label{morequeries}

Because we trained on YFCC100M~\cite{yfcc100m} with its large number of associated user tags, we found that we could query almost any word and retrieve reasonable images. Another anecdotal observation was that most images we trained on had useless tags, meaning that the noise floor was much higher than expected. Despite this, retrieving based on text queries yielded extraordinarily relevant images. 

\begin{figure}
    \centering
    \begin{subfigure}[b]{0.44\textwidth}
        \centering
        \includegraphics[width=\textwidth]{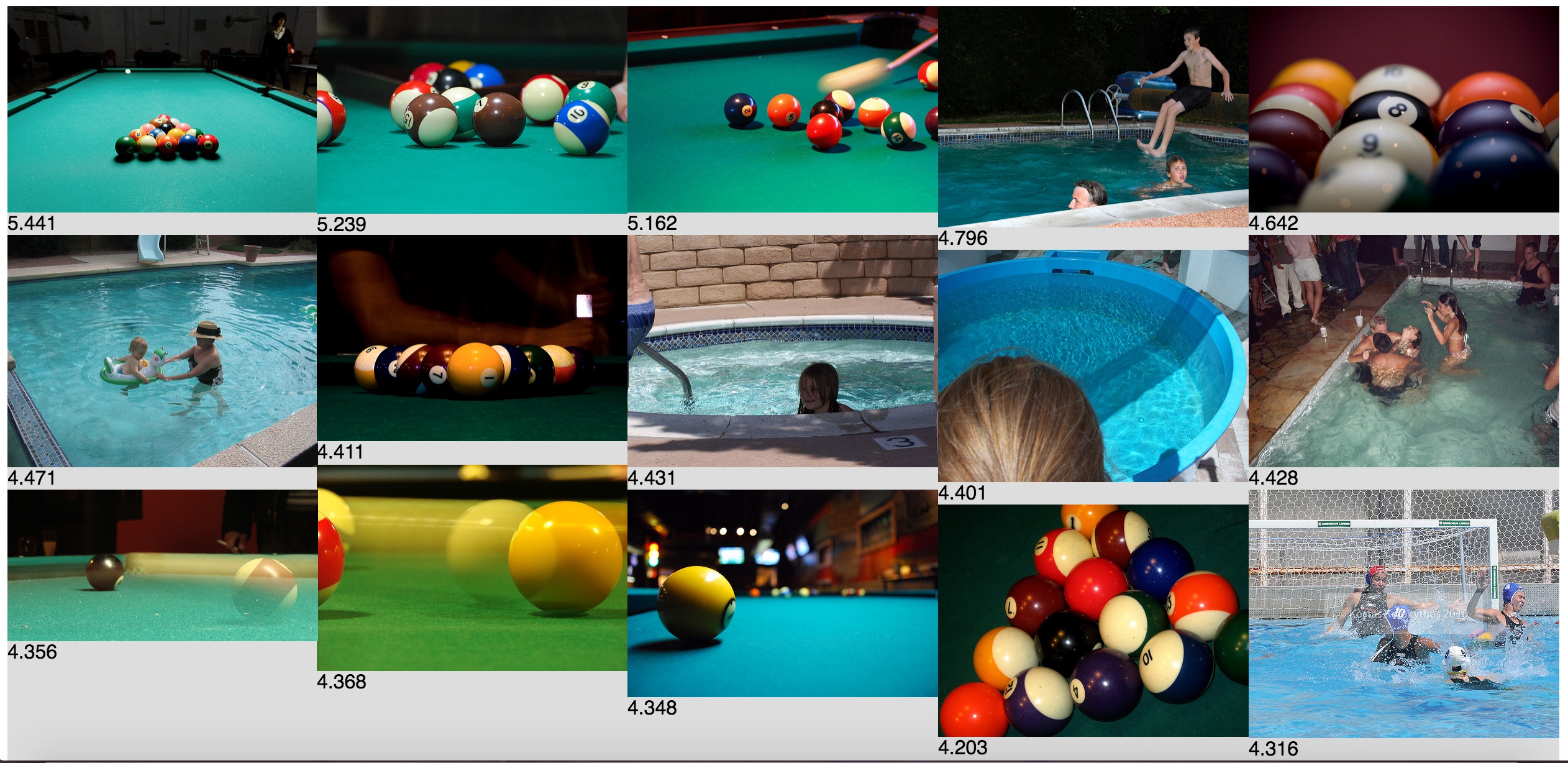}
        \caption{Retrieved images for the word query ``pool''. There are several visual representations of the word yielding pictures of both pool tables and swimming pools.}
        \label{fig:polysemy}
    \end{subfigure}
    \hfill
    \begin{subfigure}[b]{0.44\textwidth}
        \centering
        \includegraphics[width=\textwidth]{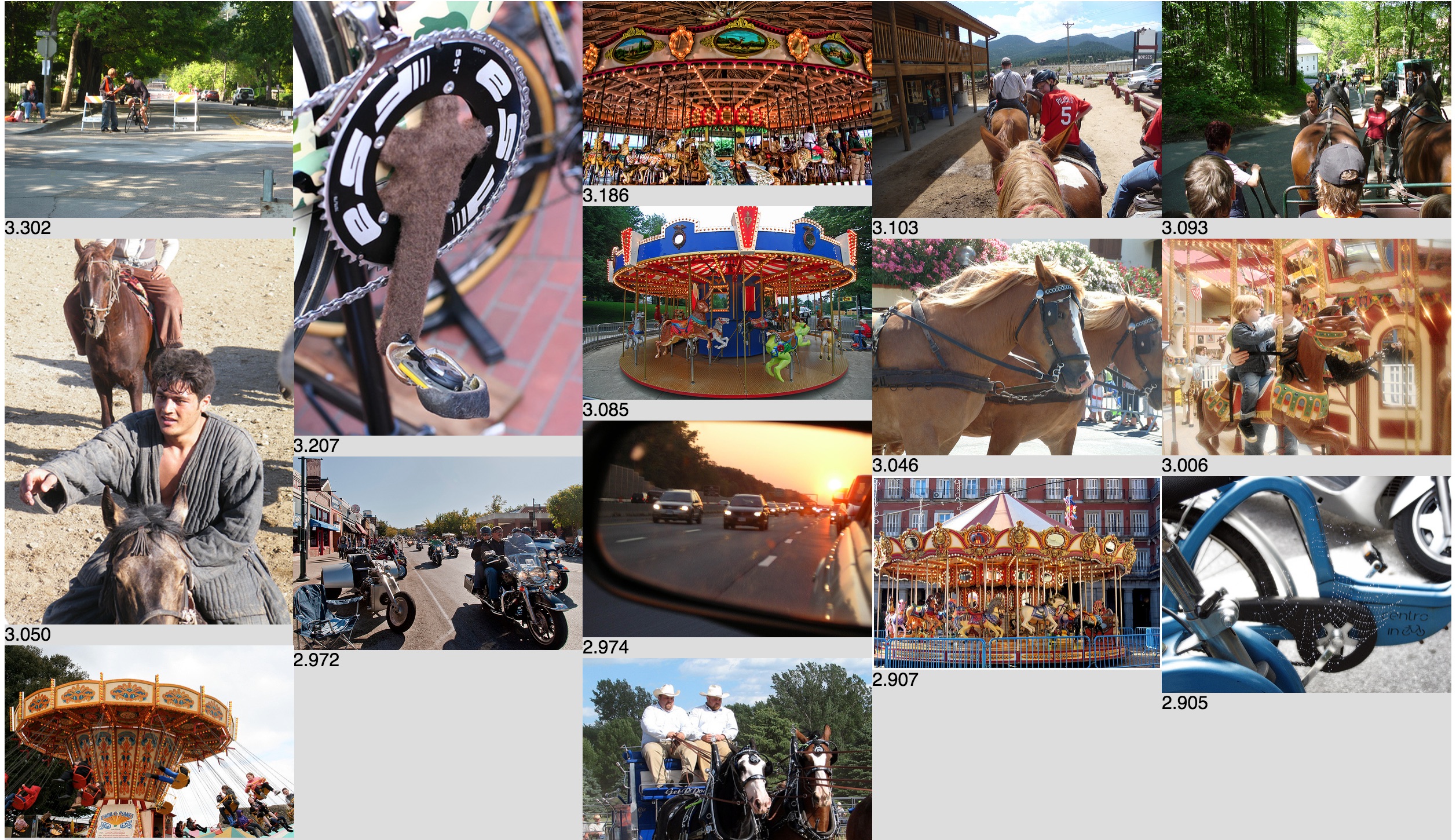}
        \caption{Retrieved images for the word query for ``ride''. Pictures returned shows that the algorithm has senses of multiple things being ridden. The noun tense of the word shows amusement park rides. The verb form shows horses, motorcycles, and bicycles.}
        \label{fig:verbs}
    \end{subfigure}
    \hfill
    \begin{subfigure}[b]{0.44\textwidth}
        \centering
        \includegraphics[width=\textwidth]{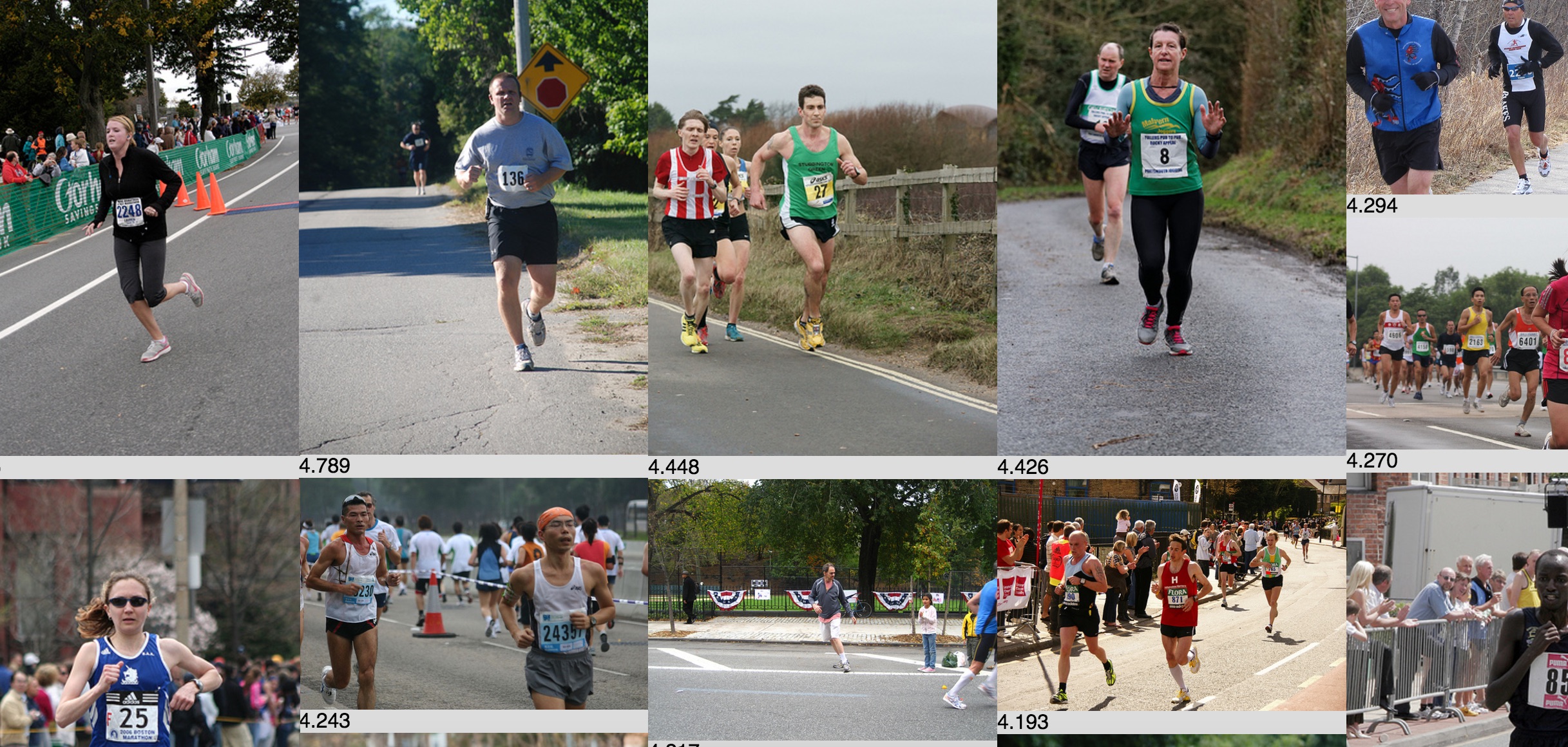}
        \caption{Retrieved images for the word query for ``running''. As expected, mostly marathon runners, but the algorithm can pick out actions in noisy metadata.}
        \label{fig:marathon}
    \end{subfigure}
    \hfill
    \begin{subfigure}[b]{0.44\textwidth}
        \centering
        \includegraphics[width=\textwidth]{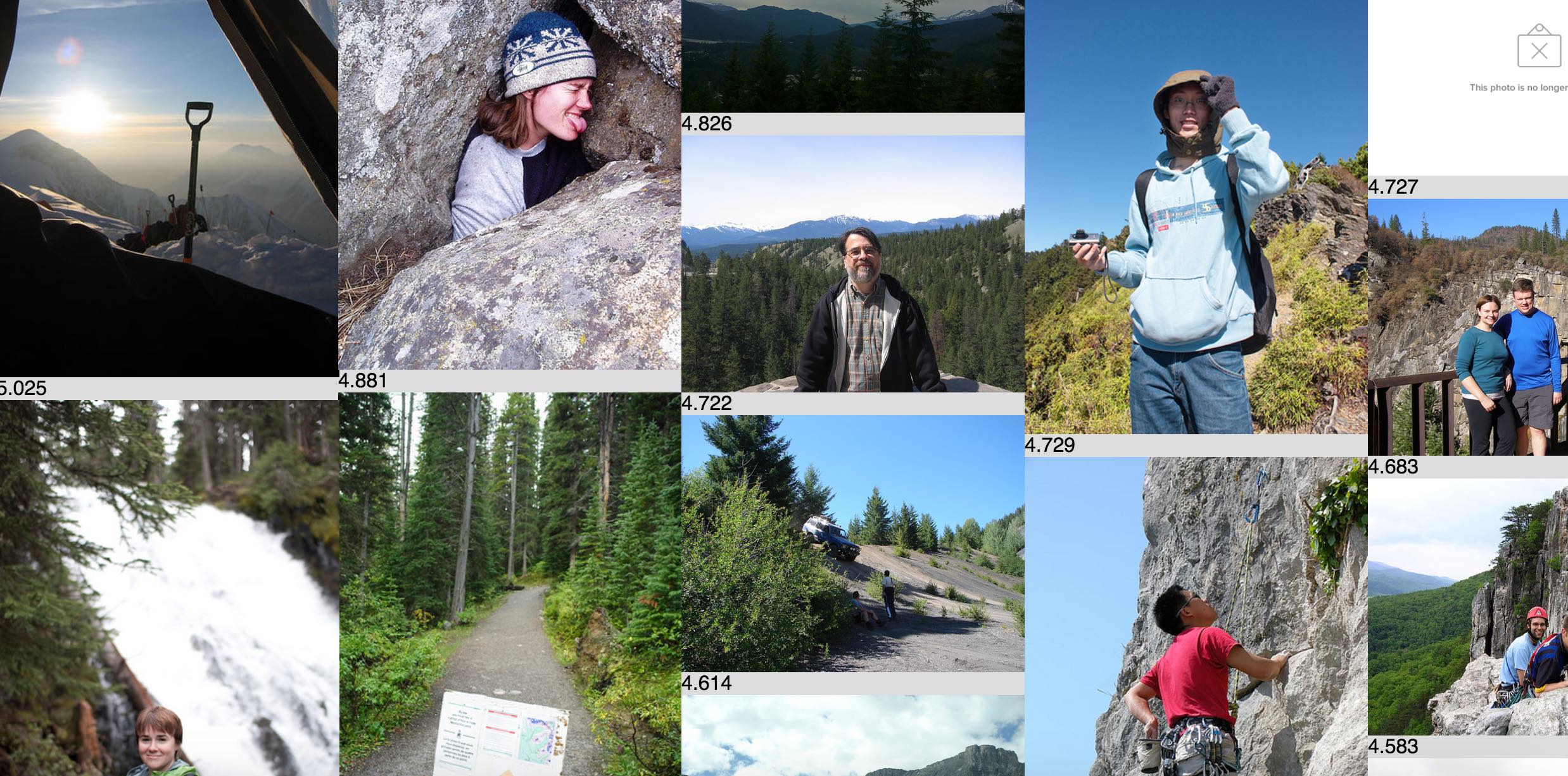}
        \caption{Retrieved images for the word query for ``rainier''. The success of this query can be attributed to people tagging which mountains they climb. These types of tags are not in datasets like Visual Genome, MS CoCo, or ImageNet.}
        \label{fig:rainier}
    \end{subfigure}
    \caption{Additonal Image Queries.}
    \label{fig:three graphs}
\end{figure}

The following qualitative examples are the more difficult queries demonstrating specific corner cases that we tested. Fig.~\ref{fig:polysemy} is a specific case of polysemy, where two senses of the noun ``pool'' are tested, and the image query retrieves both. Fig.~\ref{fig:verbs} is not polysemous in nature, but demonstrates the algorithm's ability to retrieve images associated with actions as well as the objects that accompany the actions. That is, even though the verb ``ride'' has a specific definition, someone can ``ride'' many things. In this case, someone can ride bicycles, carousels, and horses, and both are retrieved as relevant images to the query despite looking totally different.

Most datasets exclude the use of proper nouns for the reason that they do not generalize. In realistic scenarios of user generated content, though, a significant portion of metadata in images are of the names of people. Fig.~\ref{fig:rainier} is an example of querying using a proper noun. There are more difficult queries that include proper nouns (like Mt. Rainier) that are not likely to be in datasets. The returned images are typical mountain pictures (that may or may not be at Mt. Rainier.)

\section{On Evaluation Metrics}
\label{eval}

In performing evaluation, we discovered that different works used different definitions of various performance metrics. In cases where there was ambiguity, we verified with the appropriate metrics. Specifically, those used in~\cite{Guillaumin2009,fasttag} report on an average performance per image basis, and those used in~\cite{fast0tag} reported higher accuracy on a per corpus basis. For example, for precision,~\cite{Guillaumin2009,fasttag} defined precision as:
\begin{equation}
    P = \frac{1}{N}\sum_i^N \frac{TP_i}{TP_i + FP_i}. \label{right}
\end{equation}
Meanwhile,~\cite{fast0tag} definited it as:
\begin{equation}
    P = \frac{ \sum_i TP_i }{\sum_i TP_i + FP_i }. \label{eq:wrong}
\end{equation}
Here, $TP_i$ and $FP_i$ are true and false positives for image $i$. The same differences apply to recall and F-1 scores. We report with the former, \eqref{right}.

\section{Additional Results}

We include additional results of training the proposed algorithm. The datasets that we train the algorithm on in this paper supplement include Microsoft's CoCo Object Dataset~\cite{mscoco} as well as the Visual Genome~\cite{visgenome}. We do not utilize the localization information, except for our comparison with DenseCap~\cite{densecap}. Doing so would likely improve the proposed algorithm. We only run sampled versions of our code on Visual Genome, considering runtime and complexity.

Like the original paper, ``XEntropy'' stands for cross-entropy using fixed word vector, and ``Optimized'' stands for optimized word vectors. ``Average WordVec'' means that we have taken the average of vectors representing words in the metadata and used that as the target. We also trained a variant of DenseCap~\cite{densecap}, called ``DenseCap-objects'' to identify objects rather than phrases of scenes. This was done by taking off the recurrent network, and using multi-hot encodings as the final layer after localization.  With ``DenseCap-objects'', we do full backpropagation.

\begin{table}[h!]
    \begin{center}
     \begin{tabular}{|c | c c c | } 
     \hline
     \multicolumn{4}{|c|}{MS CoCo~\cite{mscoco}} \\
     \hline
     & \multicolumn{3}{|c|}{Top 5 Perforamnce} \\
     \hline     
     Top 5 & P & R & F1 \\
     \hline
     \hline
    Average WordVec          & \textbf{43.2} & 61.8 & 44.0 \\ 
    Correlated WV~\cite{ghetto-wdv}       & 38.0 & 70.6 & 48.4 \\ 
    Fast0Tag & 40.3 & 68.7 & 50.8 \\
    XEntropy & 40.0 & 68.6 & 50.5 \\
    Optimized & 42.1 & 68.6 & 52.1 \\
    Sampled Fast0Tag & 38.1 & 69.6 & 49.2 \\ 
    Sampled XEntropy         & 38.0 & 69.6 & 49.2 \\ 
    Sampled Optimized        & 42.0 & \textbf{72.5} & \textbf{53.2} \\ 
     [1ex] 
     \hline
    \end{tabular}
    \caption{Results on the Microsoft COCO objects dataset~\cite{mscoco}, without being trained on localization information.}
    \label{visualgenome}
    \end{center}
\end{table}

\begin{table}[h!]
    \begin{center}
     \begin{tabular}{|c | c c c | } 
     \hline
     \multicolumn{4}{|c|}{Visual Genome} \\
     \hline
     & \multicolumn{3}{|c|}{Top 5 Perforamnce} \\
     \hline     
     Top 5 & P & R & F1 \\
     \hline
     \hline
     Average WordVec & 14.2 & 4.4 & 6.7 \\ 
     DenseCap-Objects & 13.3 & \textbf{14.9} & \textbf{14.0} \\
     Sampled Fast0Tag & 14.2 & 5.6 & 8.1 \\
     Sampled XEntropy & 14.5 & 5.3 & 7.7 \\
     Sampled Optimized  & \textbf{17.6} & 10.3 & 13.0 \\
     [1ex] 
     \hline
    \end{tabular}
    \caption{Sampled methods on the Visual Genome corpus, without being trained on localization information.}
    \label{visualgenome}
    \end{center}
\end{table}

% BiBTeX files (here: refs). The IEEEbib.bst bibliography

\bibliographystyle{ieee}
\bibliography{refs}